\documentclass[conference]{IEEEtran}
\IEEEoverridecommandlockouts
\usepackage{cite}
\usepackage{amsmath,amssymb,amsfonts}
\usepackage{algorithmic}
\usepackage{graphicx}
\usepackage{textcomp}
\usepackage{xcolor}
\usepackage{hyperref}
\usepackage{booktabs}
\def\BibTeX{{\rm B\kern-.05em{\sc i\kern-.025em b}\kern-.08em
    T\kern-.1667em\lower.7ex\hbox{E}\kern-.125emX}}

\makeatletter \newcommand{\linebreakand}{ \end{@IEEEauthorhalign} \hfill\mbox{}\par \mbox{}\hfill\begin{@IEEEauthorhalign} } \makeatother

\begin{document}

\title    {Improving Collaborative Storytelling with a Multi-Agent Framework Based on Large Language Models}

\author{\IEEEauthorblockN{Arturo Valdivia}
\IEEEauthorblockA{\emph{Data Science Section} \\
{IT University of Copenhagen}\\
Copenhagen, Denmark \\
arturo@valdivia.xyz}
\and
\IEEEauthorblockN{Paolo Burelli}
\IEEEauthorblockA{\emph{Play, Culture and AI Section - brAIn lab} \\
IT University of Copenhagen\\
Copenhagen, Denmark \\
pabu@itu.dk}}


\maketitle

\begin{abstract}
The topic of Co-creation, \textit{i.e.}, AI agents interacting with humans to generate outputs (\textit{e.g.}, art), has gained significant attention recently. However, most studies focus on adult-human interactions in a digital setting. This paper explores a novel ludic co-creation scenario involving children and Large Language Models (LLMs) interacting through a physical board game to create written stories. Our goal is to develop a multi-agent framework capable of producing high-quality narratives suitable for young players.
At the core of our approach is an iterative Writer–Editor process in which one LLM generates stories while another evaluates them and provides feedback for refinement. Through a simulation study involving multiple LLMs, we show that this iterative interaction consistently improves the perceived quality of generated stories across successive loops. The results indicate that a small number of refinement steps may be sufficient to achieve high-quality outputs in interactive storytelling systems.
\end{abstract}

\section{Introduction}
Recent advances in artificial intelligence (AI), particularly in the field of Large Language Models (LLMs), have sparked a growing interest in applications designed to foster co-creation \cite{yannakakis2014mixed, MaherGKD18}. These applications enable AI agents to collaborate with humans to produce creative outputs such as visual art or stories. Most existing research in this area focuses on digital settings (\textit{e.g.}, mobile apps or screen-based devices) and targets adult users. In contrast, this paper explores a novel case involving young children aged three to six years, who interact with AI through a physical, screenless game board called YOLI.

\begin{figure}[!ht]
\centering
\includegraphics[width=0.45\textwidth]{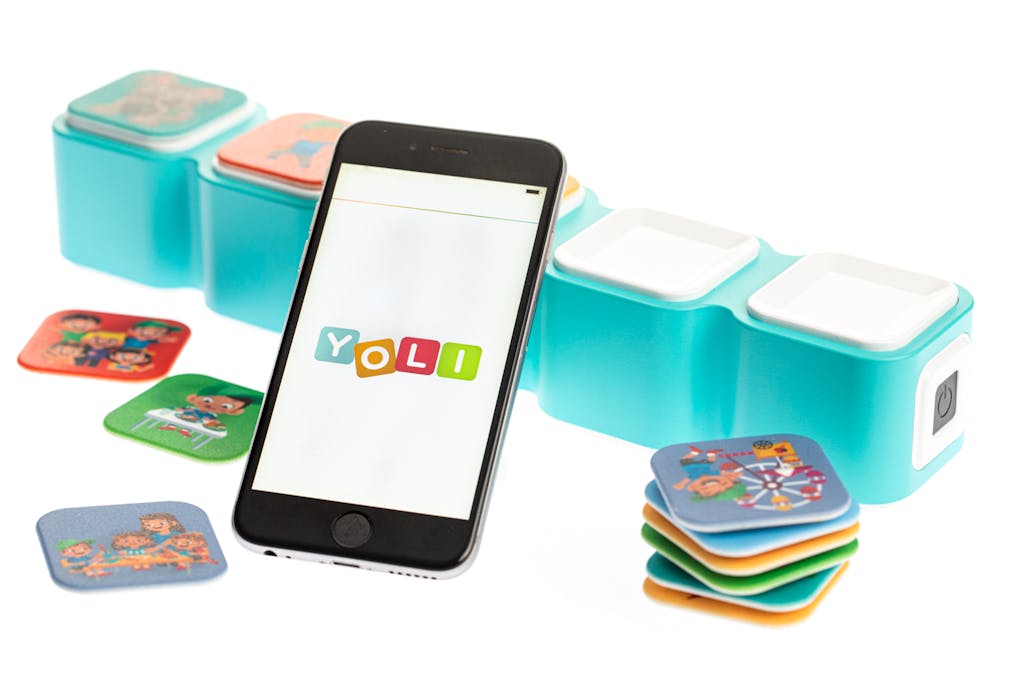}
\caption{Yoli: A physical board game for young children. Different tiles are selected sequentially to achieve a goal. The board is connected to an app that is used by the teacher who supervises the game session. The app provides AI capabilities to the board.}
\label{YOLI}
\end{figure}

The \textit{YOLI board} (see Fig. \ref{YOLI}) enables players to select various elements for their story, such as the main characters or the setting. These elements are drawn from a large pool of options, resulting in a vast number of possible combinations for players to explore during play sessions. This story co-creation problem closely resembles the challenge of open-endedness in \textit{Procedural Content Generation} (PCG) \cite{shaker2016procedural}, where research focuses on algorithms capable of producing infinite innovation \cite{mariogpt}. Similarly to PCG, it is crucial in this context to balance the generation of diverse content with ensuring the quality and appropriateness of that content. This is especially important because YOLI’s target audience is young children, making the quality and appropriateness of generated stories paramount.

To evaluate the quality and appropriateness of the generated stories, the existing literature (see \cite{ValdiviaBurelli2025, YANG_JIN, HERRERA} and references therein) suggests a wide array of criteria, including \textit{Relevance}, \textit{Diversity}, \textit{Fluency}, \textit{Coherence}, \textit{Completeness}, \textit{Clarity}, \textit{Commonsense}, \textit{Informativeness}, \textit{Character Development}, \textit{Interestingness}, \textit{Empathy}, and \textit{Surprise}. We refer readers to Fig. 2 in \cite{ValdiviaBurelli2025} for concise definitions of these dimensions.

In this paper, we focus on the \textit{Relevance} dimension, which assesses whether the story aligns and accurately incorporates the inputs provided by the players. This aspect of story quality closely relates to the problem of \textit{hallucinations} in LLMs. As defined by \cite{HUANG}, hallucinations can be categorised into two types:  \textit{Factual hallucinations}, referring to discrepancies between generated content and verifiable real-world facts; and  \textit{Faithfulness hallucinations}, describing deviations from user instructions, input context, or internal consistency within the content. Avoiding faithfulness hallucinations is crucial for ensuring that the generated story reliably reflects the story elements provided as input by the children. Failure to do so can lead to frustration and disengagement, as documented in studies on AI-based player assistants~\cite{Gallotta, Zhou}.

To address this issue, we propose an iterative \textit{Writer-Editor} process, in which different LLMs function as Writers that generate stories and Editors that evaluate and provide feedback on these stories. The Editors assign a numerical score, interpreted as the perceived quality of the story. We demonstrate that the perceived quality of generated stories improves after successive iterations of the Writer-Editor loop. Our numerical study suggests that this improvement may converge after just a few number of iterations, and gives evidence suggesting that this speed of convergence may be directly proportional to the sophistication of the Editor LLM.

\section{Related Works}

Prior narrative co-creation systems, such as Alvarez’s Story Designer~\cite{alvarez2022story}, relied on rule-based story grammars and mixed-initiative interfaces to guide structured plot construction. While these tools offer precise constraint handling, they require extensive manual authoring and lack linguistic flexibility for real-time generation. LLMs address this gap by enabling zero-shot, fluent story creation that dynamically adapts to diverse tile combinations. Although traditional methods could theoretically manage YOLI’s fixed role structure, LLMs significantly reduce authoring overhead while supporting open-ended variation—a practical advantage for scalable, child-centered interactive storytelling.

The approach of employing multiple LLMs with a variety of roles has recently been applied~\cite{Dong} in the context of prompt optimization. The authors explore how a mixture of LLMs playing the role of an \textit{actor} and a \textit{critic} can offer a potential solution to the situations where LLMs struggle to capture user intent from fuzzy, complex, or \textit{ low-quality} prompts written by humans. Their proposed iterative process of execution and evaluation guides the LLM in refining the prompt to better align with the desired task and improve performance. The study's results suggest that this multi-agent approach elevates the relative performance of medium- and low-quality human-written prompts to a comparable level with that of high-quality human-written prompts. 

Although our approach shares similarities with that of Dong et al.~\cite{Dong}, the key distinction lies in the objective: their iterative process focuses on improving the prompt itself, whereas our iterative method is designed to increase the quality of the output generated by our system (\textit{i.e.}, the story).

A different approach to ensure the quality of LLM's outputs is based on the so-called \textit{Reinforcement Learning from Human Feedback} (RLHF) method, where a reward model is trained on human-annotated data and then utilized to fine-tune the LLM via reinforcement learning methods like \textit{Proximal Policy Optimization} (PPO) \cite{Christiano, Ziegler, stiennon2020summarize, ouyang2022training, schulman2017ppo, Bai2022a}.  

Despite these advances in the field of RLHF, a growing number of authors have highlighted that obtaining high-quality human preference data remains a significant challenge. Such data is expensive, labour-intensive to produce, and often lacks diversity, with expert annotations being particularly resource-intensive \cite{kopf2024openassistant, xu2023wizardlm, sun2024principle, peng2023instruction}. 

In an attempt to reduce the required high-quality human-annotated data for RLHF, some recent works \cite{Huang2024selfevolved, Bai2022b, lee2023rlaif} explore alternative ways to leverage the LLMs themselves, for instance, by using the feedback generated by AI to train reward models, leading to approaches typically referred to as \textit{Reinforcement Learning from AI Feedback} (RLAIF). Other works like \cite{pang2023selfimprovement, kumar2024selfcorrect, huang2022selfimprove} discuss methods of self-correction and improvement using self-generated data. As we discuss in the following sections, our approach is also based on the hypothesis that the system can rely on its own intermediate steps to iteratively improve the quality of their ultimate output.

\section{The Narrative Game Loop}

The board game employed in the research conducted in this paper is called YOLI. More specifically, YOLI is a digital-physical platform that supports a collection of games —or, \textit{YOLI games}— designed for young children between the ages of three and six. These games are played on the so-called YOLI board, a physical board shaped as a 5x1 grid where players can place up to five different tiles from a tile box –see Figure \ref{YOLI}. These games are designed in collaboration with educators and cover the six curriculum themes outlined by Denmark's Ministry of Children and Education \cite{Ministry}: Versatile personal development; Social development; Language and communication; Body, senses and movement; Culture and community; and Nature, outdoor life and science.

While playing with YOLI, children can discuss the pieces with each other, and the interaction with the board motivates them to try out different options. The board provides feedback to the children in the form of vibrations and noise that help them understand whether the tile they have placed is valid or invalid, according to the specifications of the particular game they are playing.

One of these games in particular revolves around guiding the children through a sequential selection of tiles that later on are used to create a story, which is generated with an LLM and then read out loud to the children using another text-to-speech agent. After the story ends, the YOLI board ejects all current tiles and the game starts once again from the beginning.

This particular game is the focus of our research. In what follows, we describe two applications of LLMs in the game: The first one is related to guiding the children through the gameplay, which turns out to be a novel way to understand Dynamic Difficulty Adjustments \cite{xue2017DDA} for young players. The second focuses on the application of LLMs to create high-quality stories for these young players. 


\subsection{Guiding players throughout the tiles selection}
The narrative framework underlying the generated stories is structured around the following key elements:

\begin{itemize}
\item \textbf{Protagonist}: The central character of the story. A girl, a firefighter, a cat, a teacher are examples of tiles that can be provided in the tiles box, and the children can select any of these as the protagonist for the story. 
\item \textbf{Location}: The setting where the story takes place, \textit{e.g.}, a kitchen in a private home, a playground.
\item \textbf{Mood}: An emotional state (\textit{e.g.}, happiness, amusement) or atmosphere that the protagonist experiences, influencing their actions and interactions.
\item \textbf{Important Object}: An object (\textit{e.g.}, a black dress, a heart-shaped soap) of significance to the protagonist, playing a pivotal role in the story.
\item \textbf{Activity}: An action or pursuit undertaken by the protagonist.
\item \textbf{Special Appearance}: A surprise element, such as a unique character (\textit{e.g.}, Santa Claus) or event, introduced by the AI agent to add an element of novelty or excitement. In particular, this element is not chosen by the players but is generated autonomously by the AI.
\end{itemize}

This induces a tile selection process that starts by selecting a character that could be the centre of the story (\textit{e.g.}, a cat). After this, the children then need to select a location for the story (\textit{e.g.}, a kitchen in a private home), a relevant object (\textit{e.g.}, a black dress), an emotion (\textit{e.g.}, happiness), etc. Once all tile positions have been filled on the board with these selections, the board connects to a web service where the story generation LLM agent is running. 

At the beginning of the game, an LLM agent facilitates the storytelling process by guiding the children through the initial stages of tile selection. The agent may motivate the children with open-ended instructions, such as: \textit{“Look at all the tiles! Pick the one you like the most. Which one looks the most fun or exciting?”}. To encourage participation and help children engage in the task, the LLM agent can provide additional cues if the selection process takes too long. For example: \textit{“Imagine the card you picked is the star of the story. What would they do?”}; or \textit{“Close your eyes and picture them doing something amazing, like flying, finding treasure, or helping a friend. If you think they’d be great in your story, that’s your hero!”}. These hints are designed to foster creativity and active participation, ensuring that the game remains dynamic and interactive.

 

    

\subsection{Co-creating high-quality stories via Writer-Editor loops}

Once the children have selected all the tiles with key story elements, the YOLI board sends these tiles as input to the app connected to the board. The app, in turn, tasks a multi-agent system with generating a story that incorporates those inputs. 

The first agent involved is the LLM, which will be responsible for story generation. We refer to this agent as a \textit{Writer}. Obtaining a high-quality story is not a simple task, to improve the chances of generating a story that is suitable for the target audience, we employ another LLM agent, the \textit{Editor}, whose role is to evaluate the story generated by the \textit{Writer} and provide a critique that can be used by the \textit{Writer} to improve the story in a subsequent stage. This completes what is referred to as the first \textit{Writer-Editor loop}.

\begin{figure}[t!]
\centering
\includegraphics[width=0.45\textwidth]{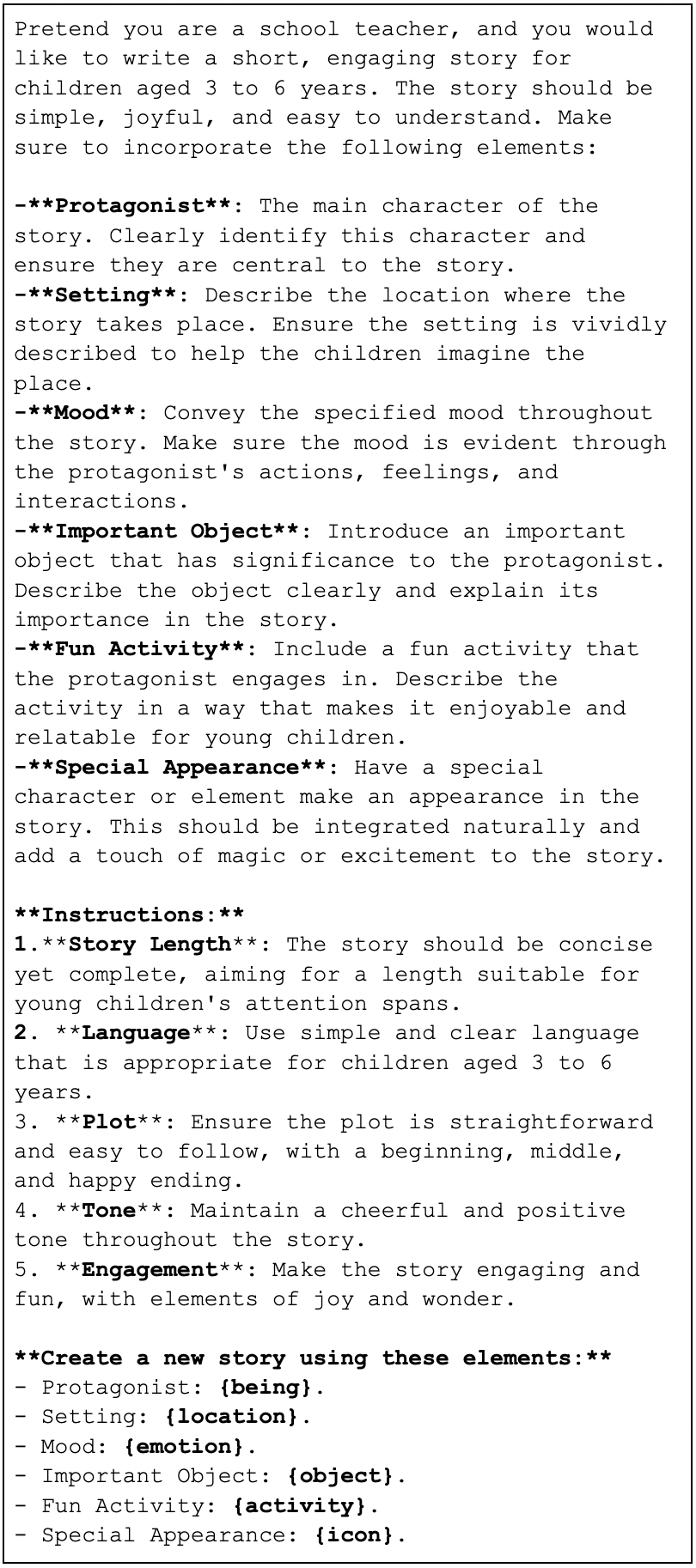}

\caption{\textbf{Writer's prompt for the first loop.} NOTE: For all prompts in this paper, some keywords have been marked in bold to improve readability. All input parameters (\textit{e.g.}, protagonist) for the prompts are enclosed between curly brackets, and these parameters are replaced by the corresponding tiles at execution time.}
\label{actor_0}
\end{figure}

To initialise the process, the Writer receives a zero-shot prompt as the template shown in (Fig. \ref{actor_0}). 

\begin{figure}[t!]
\centering
\includegraphics[width=0.45\textwidth]{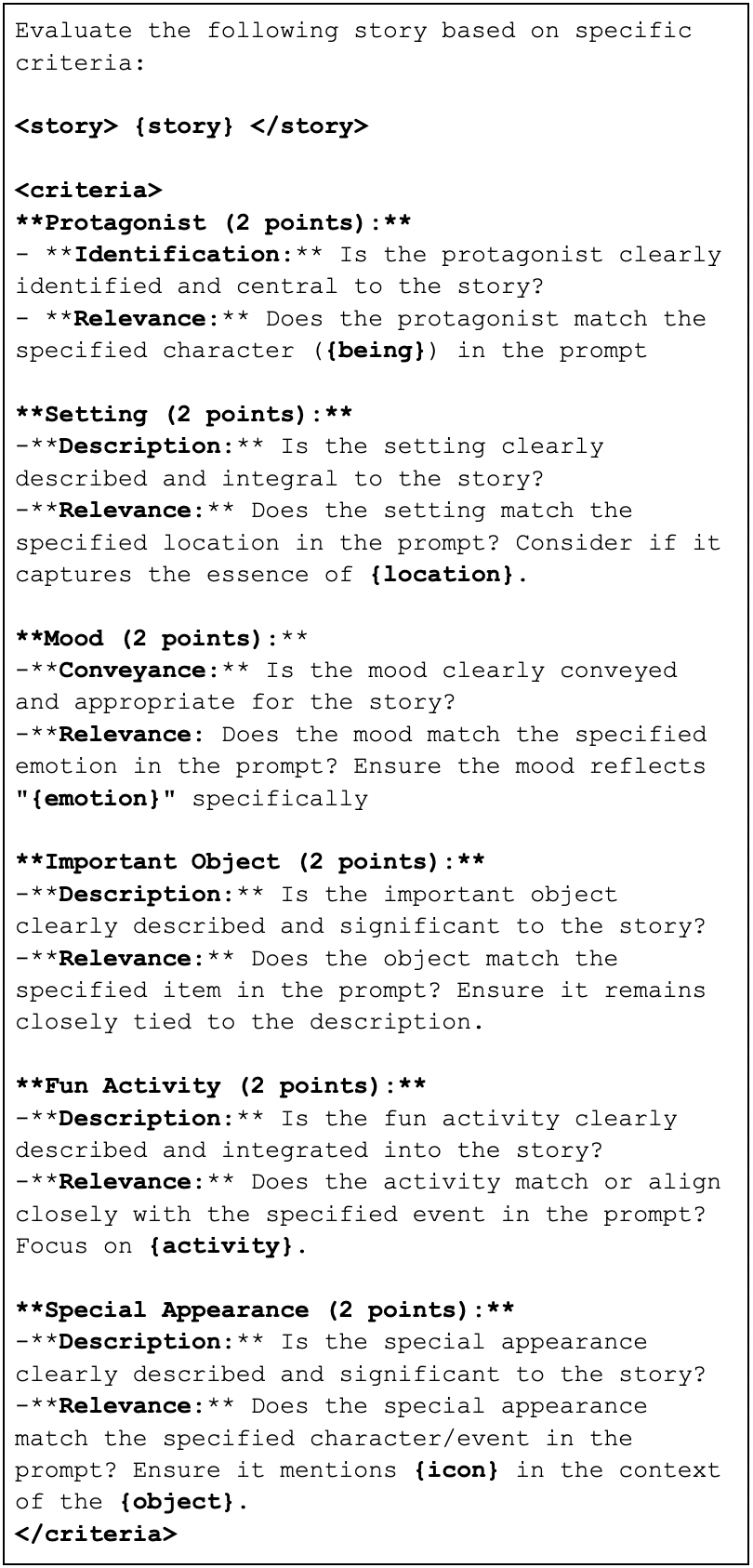}

\caption{\textbf{Editor's prompt for the first loop. Part 1/2.} 
}
\label{critic_01}
\end{figure}

Beyond the specific wording and syntax used in this prompt, what it is relevant to highlight is that it consists of four main components: A header introducing the task and personality expected; the description of the story elements and a brief explanation of how these should be used in the story; basic instructions on high-level characteristics of the story; and finally, the specific values to be used for each of the story elements. For the sake of clarity, there are no further instructions in terms of formatting the output.

\begin{figure}[h!]
\centering
\includegraphics[width=0.45\textwidth]{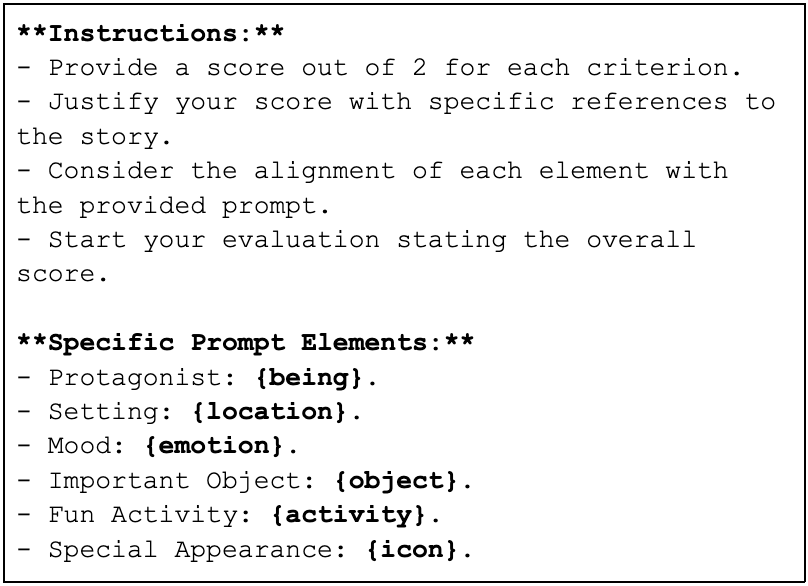}

\caption{\textbf{Editor's prompt for the first loop. Part 2/2.}}
\label{critic_02}
\end{figure}

Once the initial story is generated by the Writer, the Editor proceeds to evaluate said story using the criteria described on its rubric. The goal of this rubric (see Fig. \ref{critic_01}) is to standardise the evaluation of each single story element. In the second part of the prompt (see Fig. \ref{critic_02}), the instructions aim at ensuring that the output contains not only an overall score---which is assigned based on the provided rubric---but also the rationale behind it. This is useful to guide the Writer during the next iteration.

\begin{figure}[t!]
\centering
\includegraphics[width=0.45\textwidth]{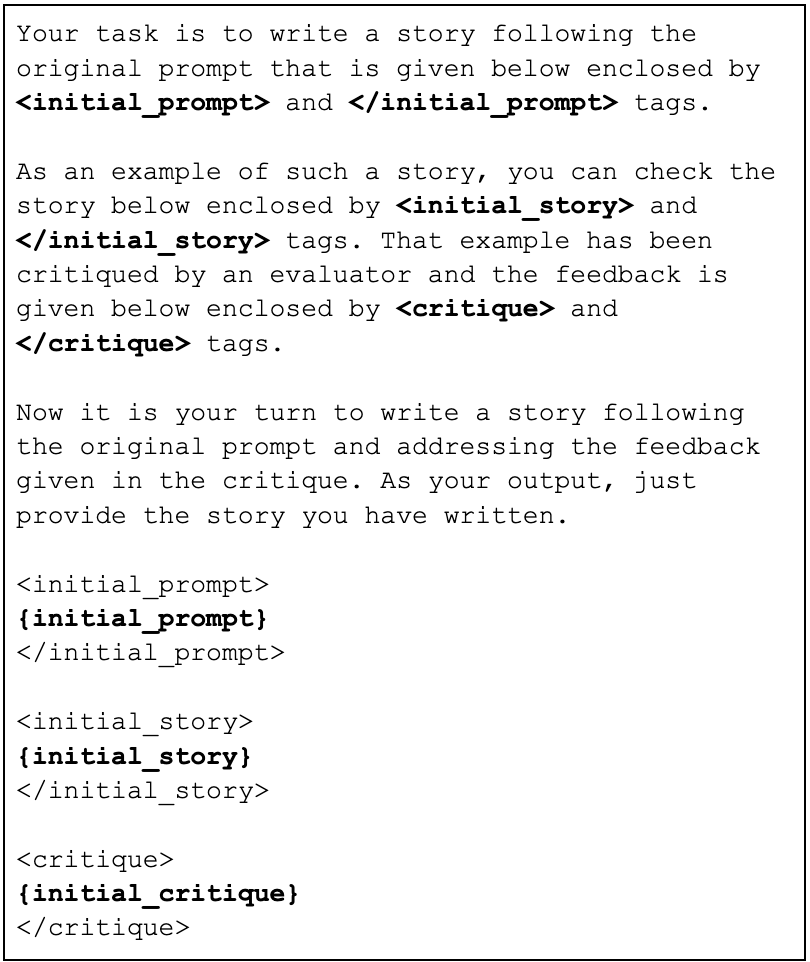}

\caption{\textbf{Writer's prompt for the subsequent loops.} 
}
\label{actor_j}
\end{figure}

After this initial loop, a new instance of the Writer agent receives a new one-shot prompt (Fig.\ref{actor_j}), which includes the zero-shot prompt given to the original Writer (Fig.\ref{actor_0}), the story generated, and the critique written by the Editor. Based on this information---containing both an overall score and the rationale behind it---the Writer aims at generating an improved version of the story that will subsequently be assessed by a new instance of the Editor. This completes the second Writer-Editor loop, and the process continues in the same manner until the predefined stopping criterion determines that the story has achieved its highest score and is ready to be shared with the children, at which point a text-to-speech service passes the text-based output from the final Writer to an audio version of the story that can be read aloud to the children. We refer to this iterative process as the \textit{Writer-Editor loops}, which is summarised in Figure \ref{Loops}.



\section{Simulation-Based Study of Writer-Editor Loops}

In this section, we conduct a simulation study to assess the behaviour of the aforementioned Writer-Editor loops. More specifically, we focus on two aspects: On the one hand, we aim to determine whether this iterative process leads to an increase in the story's quality as perceived by the LLM playing the role of an Editor.

On the other hand, we aim to estimate the speed of convergence of this method. For this purpose, we employ as the Writer an instance of the Gemma 2 LLM~\cite{gemmateam2024gemma2improvingopen} having 2 billion parameters. Playing the role of an Editor, we consider a variety of LLMs with different parameter sizes: the Gemma having 2 billion parameters, along with the 9 and 27 parameter versions; the Llama 3.1 model~\cite{dubey2024llama} having 8 billion parameters; and the Mistral model~\cite{jiang2023mistral} having 7 billion parameters. These models are run manually using the Ollama Local AI model management tool \cite{ollama}, and no particular parameters (\textit{e.g.}, temperature) are imposed at the time of inference; instead, the default values were used.

The stories we consider have a structure based on the following key elements: protagonist, location, mood, important object, activity, and special appearance. In what follows, we define as \textit{tile tuples} the choice of a specific value for each of the story's key elements. 
 

 For this simulation study, we consider a parameter space formed by roughly 1000 tile tuples. Notice that a single tuple can lead to an arbitrary number of different stories. 

\begin{figure}[ht!]
\centering
\includegraphics[width=0.45\textwidth]{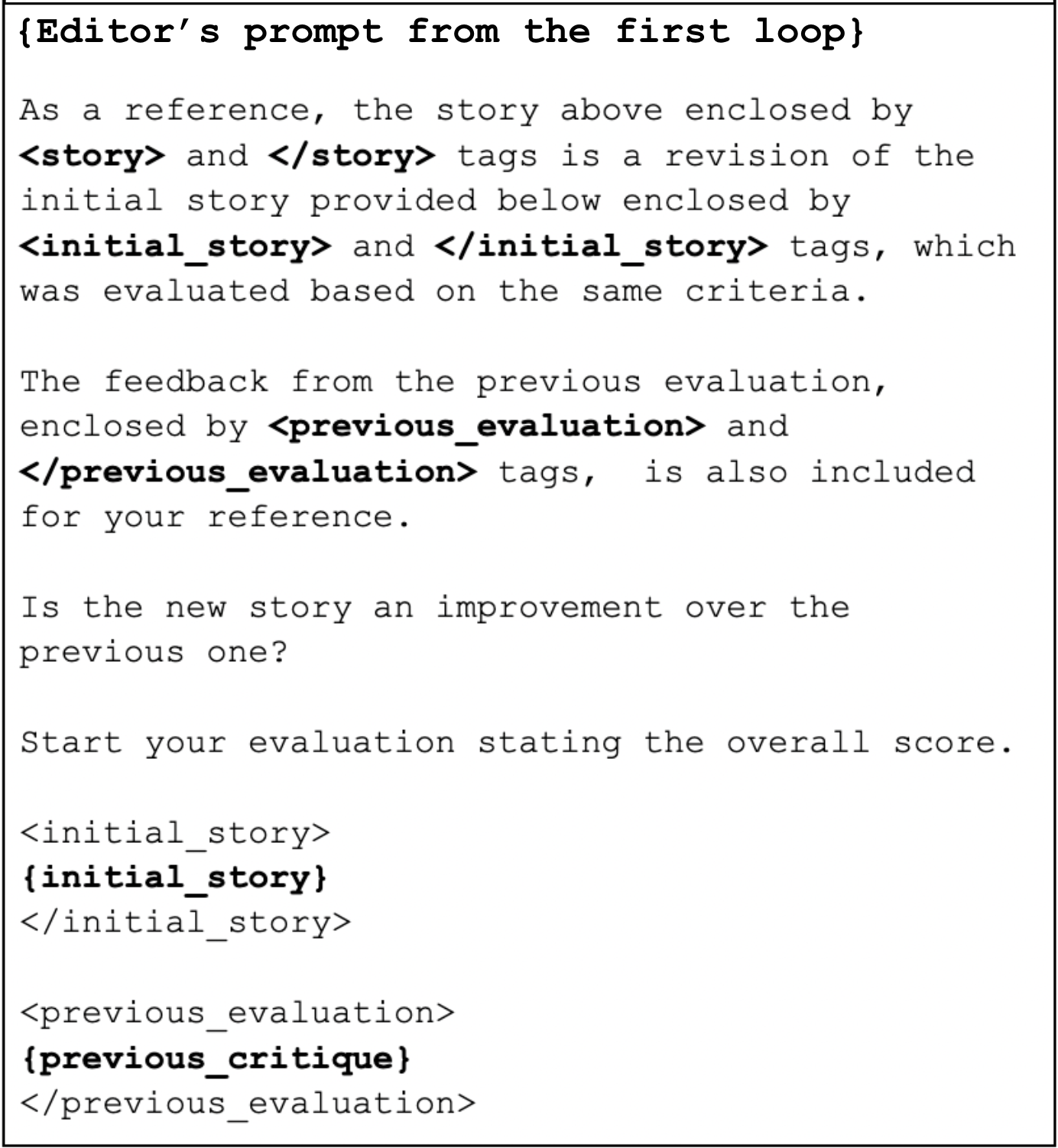}

\caption{\textbf{Editor's prompt for the subsequent loops.} Notice that this prompt starts with \texttt{{Editor’s prompt from the first loop}}. This placeholder will be replaced by the actual Editor’s prompt used in the first loop. Recall that the Editor’s prompt template from the first loop is shown in Fig. \ref{critic_01} and Fig. \ref{critic_02}. }
\label{critic_j}
\end{figure}

\begin{figure}[b!]
\centering
\includegraphics[width=0.45\textwidth]{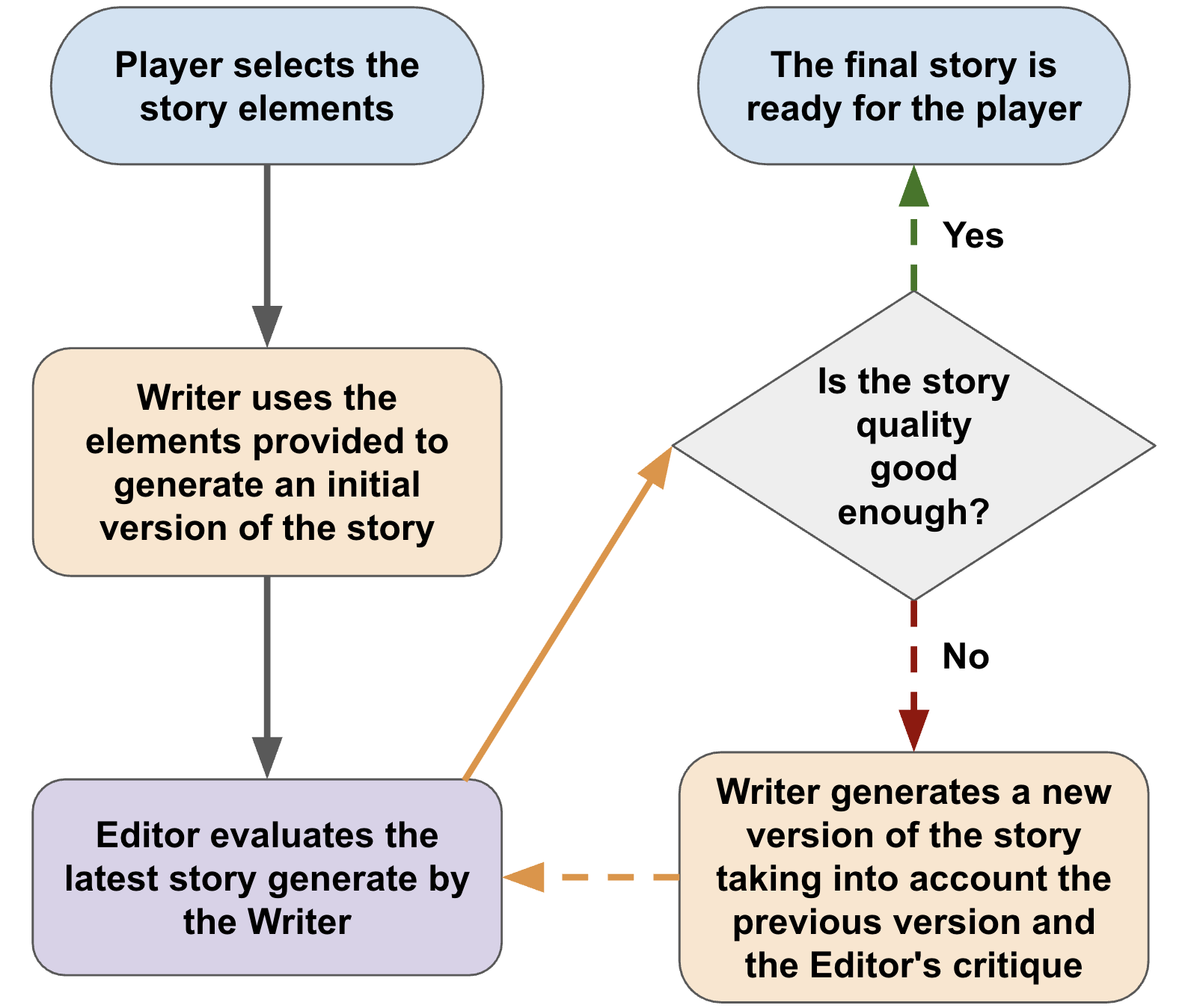}

\caption{Illustration of the Writer-Editor Loops used in the simulation study.}
\label{Loops}
\end{figure}

For each initial story written by our \textit{Writer}, we conduct five Writer-Editor loops. For the first loop, the same initial story is evaluated by each of the Editors---and thus we end up with five critiques on the same story, \textit{i.e.}, one critique for each Editor. In the second loop, the Writer generates one new story for every critique, and the new story is then evaluated by the same Editor who wrote the critique leading to this new story. The process is repeated until all loops have been completed.

The prompt for the Editor specifically requires producing a numeric score between 0\% and 100\%; we use that score to quantify the improvement in the quality of the story after each Writer-Editor loop. It is important to note that this score is not an absolute measure of quality but rather a synthetic measure of the quality as perceived by each Editor.

\subsection{Results}
Figure \ref{scores} below illustrates the average scores assigned by the Editor, as tracked cumulatively through successive iterations of the Writer-Editor loop. To provide a measure of the fluctuation of these averages, Table \ref{dispersion} shows the average score achieved along with the standard deviation observed. In the following, we describe the three key observations we derive from these data.

First, the perceived quality of the story follows an increasing trend for the five {Editors}, \textit{i.e.}, the average quality score increases after each Writer-Editor loop. This is in line with our expectations regarding the effect of the Writer-Editor loops. While an exhaustive comparison between different choices of LLMs for the Writer and the Editor agents is outside of the scope of this paper, we mark as promising the fact that this observation holds true for the five Editors considered here, independently of the model family and their number of parameters. 

Second, the biggest improvement occurs the first time the Writer creates a new story, taking into account the critique made by the Editor. That corresponds as well to the point at which we pass from the zero-shot type of prompt (see Fig. \ref{actor_0}) to the one-shot prompt (see Fig. \ref{actor_j}) that incorporates the previous story and its critique as part of the prompt. From this point onward, the improvement in story quality grows at a significantly lower rate. 

\begin{figure}[h!]
\centering
\includegraphics[width=0.45\textwidth]{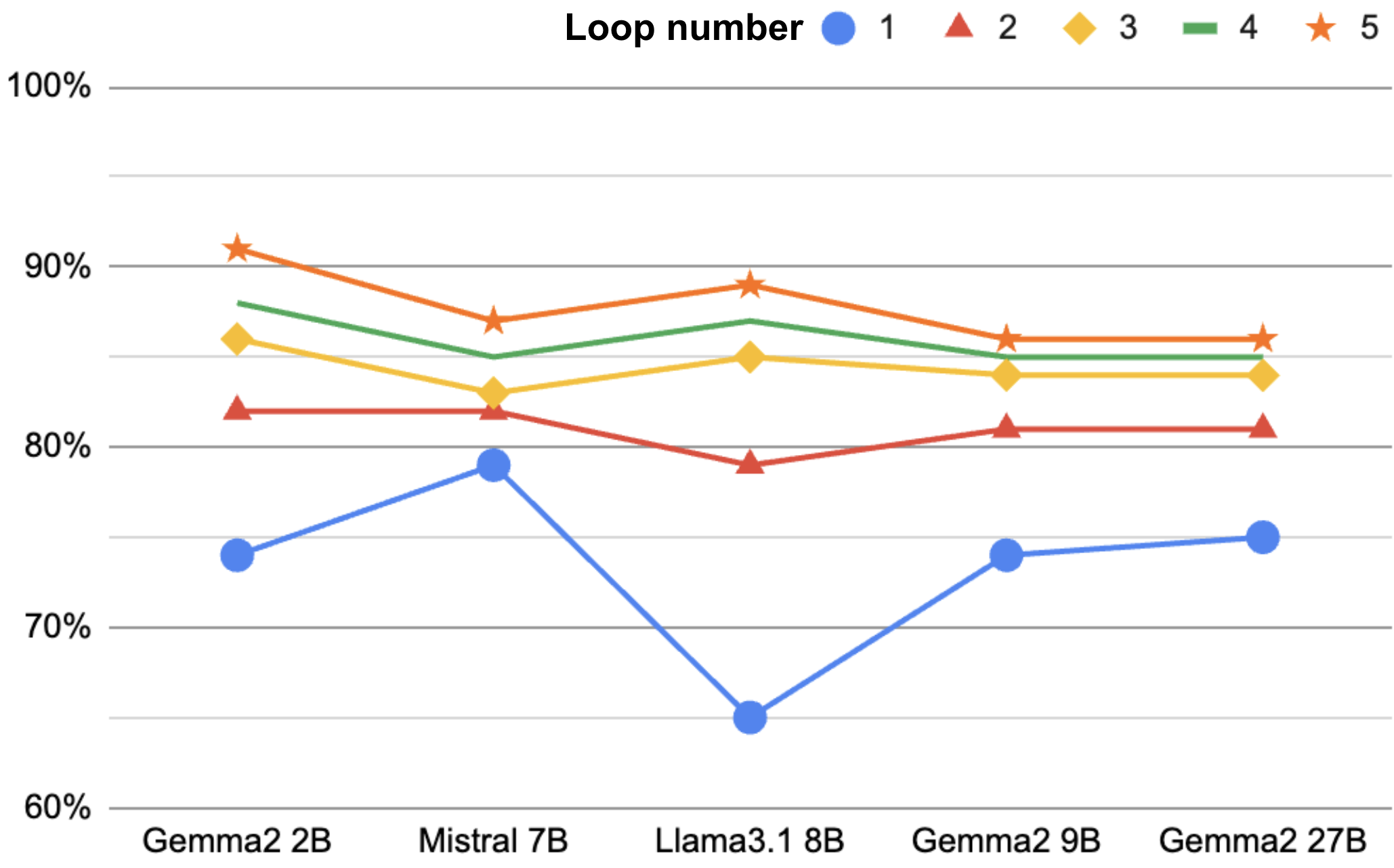}

\caption{Average story quality score after each Writer–Editor iteration. Scores increase consistently across iterations for all models used as Editors.}
\label{scores}
\end{figure}

Third, the dispersion on the scores seems to be at its lowest point, typically between the second and the third iteration. Upon inspection of some examples, it would appear that one potential explanation for this behaviour is the fact that after two or three iterations, a story can achieve such a very high score (\textit{e.g.}, 100\%) that any further attempts to improve an already highly scored story may inadvertently produce a suboptimal version.

\begin{table}[b!]
    \centering
    \begin{tabular}{c c c c c c}
Model & 1st & 2nd & 3rd & 4th & 5th\\
        \hline
        2B   & 74 $\pm$ 8  & 82 $\pm$ 7  & 86 $\pm$ 5  & 88 $\pm$ 5  & 91 $\pm$ 7 \\
        7B  & 79 $\pm$ 9  & 82 $\pm$ 8  & 83 $\pm$ 9  & 85 $\pm$ 12 & 87 $\pm$ 13 \\
        8B & 65 $\pm$ 9  & 79 $\pm$ 10 & 85 $\pm$ 9  & 87 $\pm$ 12 & 89 $\pm$ 12 \\
        9B   & 74 $\pm$ 7  & 81 $\pm$ 5  & 84 $\pm$ 6  & 85 $\pm$ 5  & 86 $\pm$ 5 \\
        27B  & 75 $\pm$ 11 & 81 $\pm$ 4  & 84 $\pm$ 10 & 85 $\pm$ 13 & 86 $\pm$ 13 \\
        \hline
    \end{tabular}
    \caption{Average score and standard deviation ($\sigma$) across five Writer–Editor loops. For brevity, the percentage symbol is omitted, and model names are abbreviated by their parameter size. For example, the first cell indicates that after the first loop, the Gemma2 2B Editor achieves an average score of 74\% with $\sigma=8\%$, signalling that scores typically range from 66\% to 82\%. }
    \label{dispersion}
\end{table}

The highest score was observed when using instances of Gemma2 2B to act both as Writer and as Editor. This may raise the concern of a \textit{self-preference bias} (\textit{cf.} \cite{wataoka2025selfpreference}), however, upon a manual inspection of a variety of examples, we believe that this phenomenon is mitigated by the use of a guided rubric for the evaluation as seen in Fig. \ref{critic_01}. Furthermore, as a contrast, it is relevant to note that when the Llama 3.1 8B model is used as Editor, we actually see a higher increase in quality scores when comparing the first versus the last loop, and the final score (\textit{i.e.}, $89\%\pm12\%$) is close in range to that of the Gemma2 2B Editor (\textit{i.e.}, $91\%\pm7\%$). The scores obtained with the Editors in the Gemma2 family showed the lowest scores after five loops.

\subsection{Optimal stopping time for the Writer-Editor loops}

As with any iterative process, it is natural to question the optimal number of iterations needed to achieve a sufficiently high score. In this study, we limited our analysis to five Writer-Editor loops. However, given the slow increase in scores observed after the third loop, we theorise that the optimal stopping point lies within these five iterations. 

\begin{figure}[h!]
\centering
\includegraphics[width=0.45\textwidth]{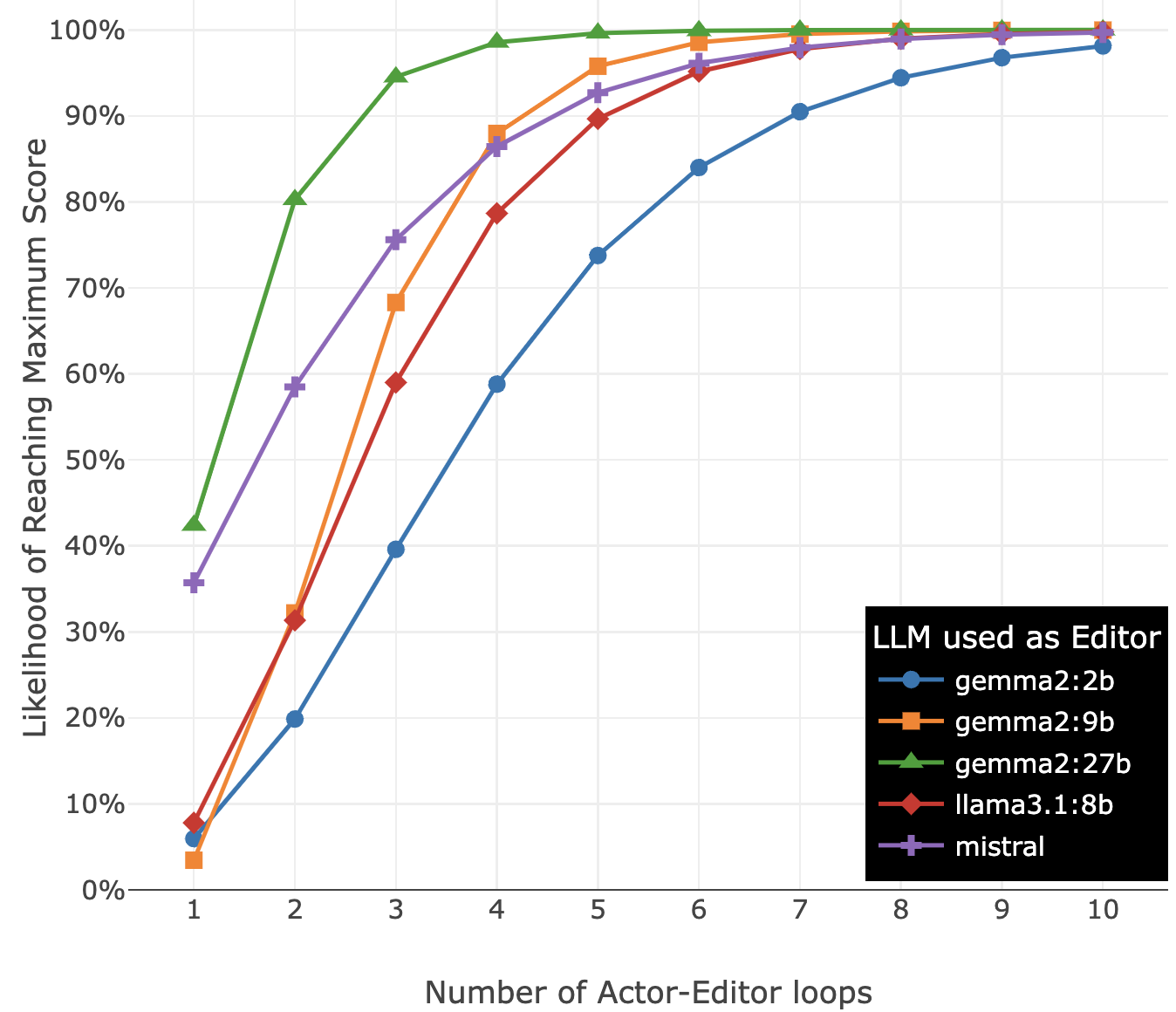}

\caption{Each line indicates how likely it is for each critic to reach its highest score as a function of the number of Writer-Editor loops. }
\label{Survival}
\end{figure}

To explore this further, we used the generated data to calibrate a classical survival model, where the \textit{mortality event} is defined as the first iteration in which the Writer-Editor loop fails to improve the score. More specifically, we used a discrete-time analogue of the continuous proportional hazards model \cite{allison1982discrete, prentice1978regression}. The corresponding survival probabilities are shown in Figure \ref{Survival}.

The survival model fitted for all \textit{Editors} demonstrated statistical significance ($p<0.05$). Analysis of the survival probabilities suggests that for Editors with 7+ billion parameters, there is a 90\% likelihood of achieving the highest score within five Writer-Editor loops. This is in line with our expectations. On the other hand, as an exception, we find the smaller model, Gemma2 2B, for which this probability drops to 74\%. It is also worth highlighting the apparent relationship between these probabilities and the number of parameters in the Editor models. More specifically, it would appear that larger models are more likely to achieve the maximum score earlier: At the extremes, the Gemma2 2B model exhibits the lowest probability, while the Gemma2 27B model shows the highest. In between, the three other models with parameters ranging from 7 to 9 billion demonstrate similar performance ---which serves also as a consistency test.


\section{Discussion}

This study examined whether iterative Writer–Editor loops can improve the quality of LLM-generated stories for young children. The findings of the simulation study provide empirical support for this approach.
Across all Editor models considered, the average quality score increased after successive iterations of the Writer–Editor loop. This suggests that critiques generated by an Editor LLM can effectively guide the Writer LLM in refining its output and addressing weaknesses in the generated narrative. Importantly, this improvement trend was observed across Editors with different parameter sizes and model families, indicating that the framework is not tied to a specific model architecture.

These findings complement existing work on exploring self-improvement and AI-generated feedback in LLM systems. In particular, our results provide empirical support for approaches based on self-critique or AI feedback, such as those explored in Reinforcement Learning from AI Feedback (RLAIF) and related self-correction methods. However, unlike approaches that rely on reinforcement learning or preference optimization, the improvements observed in our framework emerge entirely through inference-time interactions between LLM agents. Because the Writer–Editor loop does not require additional training data or model fine-tuning, it may offer a practical alternative in development contexts where collecting human preference data or training reward models is impractical.

Another notable observation concerns the speed of convergence of the iterative process. The largest improvement occurs when the Writer first incorporates the critique generated by the Editor, corresponding to the transition from a zero-shot prompt to a \textit{richer} prompt that includes both the previous story and the Editor’s feedback. Subsequent iterations produce progressively smaller improvements, suggesting diminishing returns after the second or third loop. From the perspective of game design, this behaviour is desirable: interactive storytelling systems must operate within tight time constraints, and limiting the process to a small number of iterations may already yield most of the achievable quality improvements while maintaining responsiveness during gameplay.

Our results also suggest a relationship between the capacity of the Editor model and the speed at which the system reaches its highest score. Given a fixed Writer model, larger Editor models appear more likely to reach their maximum score earlier in the iterative process, as indicated by the survival analysis. This suggests that more capable evaluation models may guide the Writer more efficiently during the refinement process. At the same time, the consistent improvement observed across all Editors indicates that the framework can remain effective even when relatively small models are used.

From a broader perspective, the Writer–Editor loop can be interpreted as a lightweight mechanism for improving constraint satisfaction in open-ended narrative generation. In procedural content generation systems for games, a common challenge is balancing the diversity of generated content with adherence to design constraints. In our case, these constraints correspond to the story elements selected by the players through the YOLI tiles. The iterative critique process provides a simple way of encouraging the Writer to remain faithful to these inputs while still generating diverse narratives.

A key limitation of the present study is that the quality scores are generated solely by LLM-based Editors, and therefore capture only a synthetic measure of quality rather than an external, objective evaluation. While these scores are useful for analysing the internal dynamics of the Writer–Editor process, human expert assessment remains essential—particularly because the system is ultimately intended for use with young children.

A preliminary qualitative assessment conducted with two professional story writers and one YOLI designer suggests that the iterative process does, in practice, improve narrative coherence and adherence to player-selected constraints. Although modest in scope, this review provides an initial indication that LLM-generated critiques align with human judgments in many cases.


A key limitation is that quality scores derive solely from LLM-based Editors, measuring internal constraint satisfaction rather than absolute narrative quality. To validate real-world impact, future work will calibrate these scores against assessments from early-childhood educators and integrate a direct rating feature for children, ensuring iterative improvements align with human-centered engagement criteria.

\section{Conclusion} 
This paper introduced and evaluated a multi‑agent Writer–Editor framework for improving LLM-generated stories in a collaborative storytelling game for young children.
The results of our simulation study suggest that the proposed Writer–Editor framework can effectively refine generated narratives, with the most significant improvements occurring within the first two to three iterations of the loop and the optimal stopping point likely within five iterations according to the survival analysis. Our findings also indicate that, given a fixed Writer model, larger Editor models tend to reach their highest score more quickly. Because the Writer–Editor process operates entirely at inference time and does not require annotated datasets or additional training, it offers a lightweight alternative to training-intensive alignment approaches such as RLHF or RLAIF. This makes the framework particularly suitable for game development contexts where rapid prototyping and limited resources are common, potentially benefiting both small studios and larger teams developing AI-assisted storytelling systems.
As LLMs increasingly support real‑time creative interactions, the Writer–Editor framework offers a promising direction for developing controllable, child‑appropriate narrative systems that balance creativity with reliability.

\bibliographystyle{IEEEtranS}
\bibliography{resources/refs}

\end{document}